\documentclass{article}
\usepackage{spconf,amsmath,graphicx, url, hyperref}
\usepackage{enumitem}
\usepackage{spconf,amsmath,graphicx}
\usepackage{booktabs,multirow}
\usepackage{xcolor}
\setlist{nosep, leftmargin=14pt}

\title{Training-time domain transfer for generator-free diagnostics on novel modalities in computational pathology}

\name{\begin{tabular}[t]{c}
Julia A. Mielcarz$^{1}$ \quad Andrea Camilloni$^{1}$ \quad Nita Mulliqi$^{1}$ \quad Jessica A. Frank$^{1, 2}$ \quad Anders Blilie$^{3}$ \\
Emilius A.M. Janssen$^{3,4,5}$ \quad Martin Eklund$^{1}$ \quad Kimmo Kartasalo$^{6,7,1,8}$
\end{tabular}}

\address{\parbox{\textwidth}{\centering
  \mbox{$^{1}$Department of Medical Epidemiology and Biostatistics, Karolinska Institutet, Sweden} \quad
  \mbox{$^{2}$Harvard Medical School, USA} \quad
  \mbox{$^{3}$Department of Pathology, Stavanger University Hospital, Norway} \quad
  \mbox{$^{4}$Department of Chemistry, Bioscience and Environmental Engineering, University of Stavanger, Norway} \quad
  \mbox{$^{5}$Institute for Biomedicine and Glycomics, Griffith University, Australia} \quad
  \mbox{$^{6}$ELLIS Institute Finland} \quad
  \mbox{$^{7}$Department of Computing, Faculty of Technology, University of Turku, Finland} \quad
  \mbox{$^{8}$Science for Life Laboratory (SciLifeLab), Karolinska Institutet, Stockholm, Sweden}}}    
\begin{document}
\AddToHookNext{shipout/foreground}{%
    \put(0.1\paperwidth,-\paperheight+1.8cm){\parbox[t]{0.8\paperwidth}{\centering\footnotesize
    This work has been submitted to the IEEE for possible publication. Copyright may be transferred without notice, after which this version may no longer be accessible.}}}

\maketitle
\begin{abstract}
Novel imaging modalities promise improved diagnostics, but AI development is constrained by scarce training data. Inference-time domain transfer can address this by mapping novel modalities to established ones, enabling reuse of existing diagnostic models. However, this introduces hallucination risks, computational overhead, and regulatory complexity.
Inspired by sim-to-real transfer, we propose training-time domain transfer, generating synthetic target-domain training data from archived routine samples to enable direct diagnosis on real inputs without inference-time generation.
Using unstained prostate histopathology as a proof-of-concept, we compare this approach with inference-time virtual staining and conventional H\&E-based cancer detection. Training-time transfer achieved higher concordance with the conventional benchmark than virtual staining for both raw (Cohen's $\kappa$=0.59 vs.\ 0.37; Gwet's AC1=0.73 vs.\ 0.29) and deparaffinized tissue ($\kappa$=0.66 vs.\ 0.61; AC1=0.73 vs.\ 0.64), supporting its potential for emerging diagnostic modalities.
\end{abstract}
\section{Introduction}
\label{sec:intro}
\begin{figure*}[t]
    \centering
    \includegraphics[width=\textwidth]{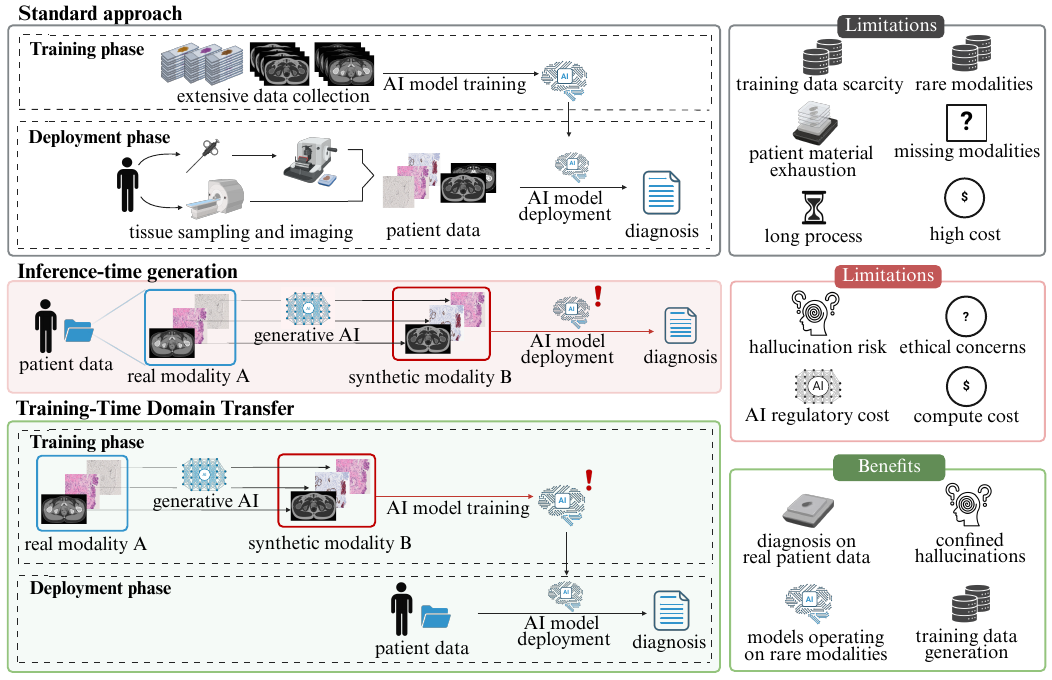}
    \caption{Comparison of conventional diagnostic AI, inference-time generation, and the proposed training-time domain transfer.}
    \label{fig:concept}
\end{figure*}

Robust diagnostic AI requires large, annotated datasets, which are scarce for emerging imaging modalities. In pathology, e.g., unstained slides, spatial multiplex proteomics, or µ-CT lack large clinical datasets, whereas archives of conventional hematoxylin and eosin (H\&E)-stained slides enable large-scale AI studies.
Computer vision addresses similar data scarcity through sim-to-real transfer, training models on synthetic data for direct deployment on real inputs, as demonstrated in autonomous driving~\cite{dosovitskiy2017carla} and robotics~\cite{tobin2017domain}. If a simulated domain, however imperfect, captures sufficient structure of the real target domain, a downstream model can be trained without real data. In medical imaging, this paradigm has been used to, e.g., synthesize MRI training data from annotated CT scans for radiotherapy contouring~\cite{kieselmann2021cross}. In pathology, the dominant response has been to deploy generative AI models to synthesize data.
A prominent example is virtual staining~\cite{latonen2024virtual}, enabled by image-to-image translation~\cite{isola2017pix2pix}
and explored, e.g., for prostate biopsies~\cite{rana2020jama}. 
These approaches operate at inference time, translating images from a novel modality into a familiar one so that existing diagnostic models can be applied. However, generative AI models can hallucinate, and realistic hallucinations such as an incorrect number of nuclei can mislead diagnosis~\cite{huang2025hallucination}.
Combining generative and diagnostic models also increases regulatory complexity, requiring separate validation and revalidation after updates.
We propose reversing this paradigm by restricting generative AI to training (training-time domain transfer, TTDT), following a sim-to-real approach (Fig.~\ref{fig:concept}). As a proof-of-concept, we generate synthetic unstained training images from archived H\&E slides for prostate cancer detection, while performing inference directly on real unstained tissue. To our knowledge, this approach is novel in computational pathology.

\section{Materials and methods}
\label{sec:methods}

\subsection{Dataset and preprocessing}
Prostate biopsy blocks from the STHLM3 trial~\cite{gronberg2015sthlm3} were re-sectioned and sequentially scanned at three processing stages: raw, deparaffinized (DP) and H\&E-stained. The original ISUP grades were retained as reference. The resulting stain-free histopathology (STHLM3-SFH) dataset comprises paired scans for training the generative models (250 raw/H\&E and 117 DP/H\&E slide pairs) and a test split of 79 slides from 67 patients scanned at all three stages (Table~\ref{tab:data}). WSIs were tiled (1024$\times$1024 px, 0.5\,$\mu$m/px) and to correct minor spatial shifts from sequential scanning, raw and DP images of the training set were each registered to their H\&E counterparts in Fiji by template matching~\cite{thomas2020multi} on random patch pairs.

\begin{table*}[t]
  \centering
  \caption{Patient and slide characteristics of the generative and diagnostic model datasets (N, (\%)).}
  \label{tab:data}
  \small
  \setlength{\tabcolsep}{3pt}
  \begin{tabular*}{\textwidth}{@{\extracolsep{\fill}}lccccccccc@{}}
    \toprule
    & \multicolumn{3}{c}{STHLM3-SFH} & \multicolumn{3}{c}{Diagnostic training} & \multicolumn{3}{c}{Diagnostic tuning} \\
    \cmidrule(lr){2-4}\cmidrule(lr){5-7}\cmidrule(l){8-10}
    & Train raw & Train DP & Test & STHLM3 & SUH & STG & STHLM3 & KUH & RUMC \\
    \midrule
    Patients & 250 & 117 & 67 & 2{,}711 & 710 & 70 & 24 & 73 & 72 \\
    Slides & 250 & 117 & 79 & 29{,}527 & 4{,}589 & 247 & 276 & 330 & 195 \\
    WSIs & 250 & 117 & 79 & 32{,}609 & 4{,}591 & 247 & 276 & 330 & 195 \\
    \midrule
    Age $\leq$ 49 & 0 (0.0) & 0 (0.0) & 0 (0.0) & 4 (0.1) & 13 (1.8) & 0 (0.0) & 0 (0.0) & 2 (2.7) & -- \\
    Age 50--54 & 15 (6.0) & 6 (5.1) & 8 (11.9) & 216 (8.0) & 35 (4.9) & 1 (1.4) & 1 (4.2) & 5 (6.8) & -- \\
    Age 55--59 & 26 (10.4) & 11 (9.4) & 10 (14.9) & 430 (15.9) & 94 (13.2) & 2 (2.9) & 2 (8.3) & 10 (13.7) & -- \\
    Age 60--64 & 68 (27.2) & 34 (29.1) & 14 (20.9) & 701 (25.9) & 137 (19.3) & 4 (5.7) & 8 (33.3) & 12 (16.4) & -- \\
    Age 65--69 & 134 (53.6) & 62 (53.0) & 29 (43.3) & 1{,}208 (44.6) & 191 (26.9) & 6 (8.6) & 13 (54.2) & 15 (20.5) & -- \\
    Age $\geq$ 70 & 7 (2.8) & 4 (3.4) & 6 (9.0) & 152 (5.6) & 240 (33.8) & 37 (52.9) & 0 (0.0) & 29 (39.7) & -- \\
    Unknown & 0 (0.0) & 0 (0.0) & 0 (0.0) & 0 (0.0) & 0 (0.0) & 20 (28.6) & 0 (0.0) & 0 (0.0) & 72 (100.0) \\
    \midrule
    Benign & 38 (15.2) & 26 (22.2) & 15 (19.0) & 23{,}521 (79.7) & 3{,}423 (74.6) & 1 (0.4) & 192 (69.6) & 108 (32.7) & 95 (48.7) \\
    ISUP 1 & 35 (14.0) & 7 (6.0) & 9 (11.4) & 3{,}571 (12.1) & 682 (14.9) & 1 (0.4) & 28 (10.1) & 65 (19.7) & 24 (12.3) \\
    ISUP 2 & 41 (16.4) & 12 (10.3) & 14 (17.7) & 1{,}265 (4.3) & 239 (5.2) & 1 (0.4) & 18 (6.5) & 63 (19.1) & 15 (7.7) \\
    ISUP 3 & 60 (24.0) & 39 (33.3) & 14 (17.7) & 494 (1.7) & 129 (2.8) & 2 (0.8) & 13 (4.7) & 49 (14.8) & 15 (7.7) \\
    ISUP 4 & 24 (9.6) & 17 (14.5) & 11 (13.9) & 377 (1.3) & 54 (1.2) & 32 (13.0) & 13 (4.7) & 19 (5.8) & 19 (9.7) \\
    ISUP 5 & 52 (20.8) & 16 (13.7) & 16 (20.3) & 299 (1.0) & 62 (1.4) & 210 (85.0) & 12 (4.3) & 26 (7.9) & 27 (13.8) \\
    \bottomrule
  \end{tabular*}
\end{table*}

\subsection{Generative models} 
We used Pix2Pix~\cite{isola2017pix2pix} with a UNet\#\# generator~\cite{korkmaz2023unetsharp}. Four models were trained: virtual staining (raw or DP to H\&E) and virtual unstaining (H\&E to raw or DP), on 19,831 (raw) or 9,238 (DP) paired tiles, with adversarial, L1 ($\lambda$=100) and Pearson correlation ($\gamma$=10) losses, using Adam (learning rate $2\times10^{-4}$) for up to 50 epochs, with the checkpoint chosen from the training curves and visual inspection at validation. Implementation: \url{https://github.com/JAMielcarz/training-time-domain-transfer}

\subsection{Diagnostic model} 
The model follows a prior multiple instance learning approach~\cite{mulliqi2025foundation}, adjusted to benign vs.\ malignant classification: an EfficientNet-V2-S feature extractor followed by gated attention~\cite{ilse2018abmil}. It was trained on 37,447 H\&E WSIs from three cohorts (STHLM3; Stavanger University Hospital, SUH; S:t Göran Hospital, STG), either directly (H\&E benchmark and inference-time virtual staining), or after virtual unstaining to each unstained stage (TTDT), with each model trained solely on synthetic data (Table~\ref{tab:data}). Ten cross-validation models were ensembled, and the threshold was optimized for F1 on a multi-cohort tuning set (STHLM3; Karolinska University Hospital, KUH; Radboud University Medical Center, RUMC). Diagnostic training and tuning cohorts (details in ~\cite{mulliqi2025foundation}) share no patients with the STHLM3-SFH test set.

\subsection{Evaluation} Five scenarios were compared: real H\&E (benchmark), synthetic H\&E from raw or DP tissue (virtual staining), and real raw or DP tissue (TTDT). 
Each approach was compared with the H\&E benchmark by DeLong's test~\cite{delong1988} on AUCs against the pathologist's diagnosis, and by Cohen's $\kappa$ and Gwet's AC1~\cite{gwet2008ac1} (robust to the high malignancy prevalence).

\section{Results}
\label{sec:results}
\begin{figure*}[t]
    \centering
    \includegraphics[width=\textwidth]{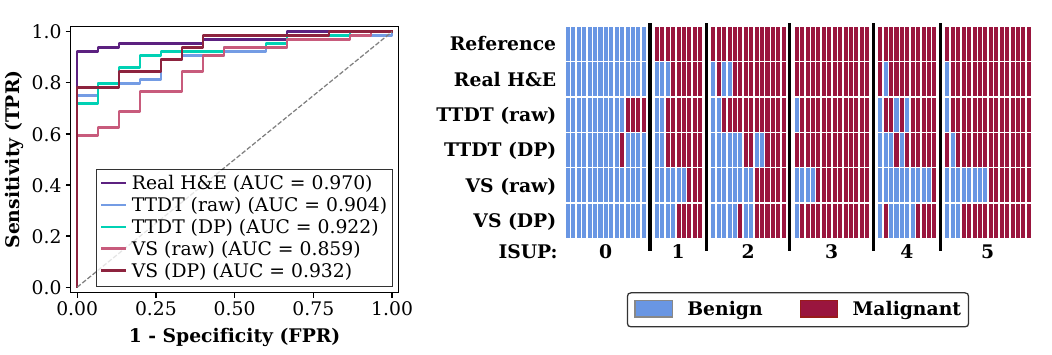}
    \caption{Test set (n=79): ROC curves (left) and slide-level predictions vs.\ the pathologist reference (right).}
    \label{fig:roc_heatmap}
\end{figure*}

Evaluated against the pathologist on the test set (n=79), real H\&E achieved the highest AUC for cancer detection (0.970; Table~\ref{tab:results}, Fig.~\ref{fig:roc_heatmap}). Virtual staining reached AUCs of 0.932 (DP) and 0.859 (raw), and TTDT reached 0.922 and 0.904. Only virtual staining from raw input had a significantly lower AUC (p=0.016) than the benchmark and a low sensitivity (0.44). TTDT showed higher concordance with the H\&E benchmark AI than virtual staining did for both raw ($\kappa$=0.59 vs.\ 0.37; AC1=0.73 vs.\ 0.29) and DP input ($\kappa$=0.66 vs.\ 0.61; AC1=0.73 vs.\ 0.64).

\begin{table}[t]
  \centering
  \caption{Diagnostic performance against pathologist labels and agreement with the H\&E model ($\kappa$, AC1). DeLong's p-values compare AUCs with real H\&E. VS: virtual staining; TTDT: training-time domain transfer.}
  \label{tab:results}
  \small
  \setlength{\tabcolsep}{2.2pt}
  \begin{tabular}{@{}lccccccc@{}}
    \toprule
    Approach & AUC & F1 & Sens. & Spec. & $p$ & $\kappa$ & AC1 \\
    \midrule
    Real H\&E   & 0.970 & 0.933 & 0.875 & 1.000 & --   & --  & -- \\
    \midrule
    VS (raw)    & 0.859 & 0.609 & 0.438 & 1.000 & 0.016 & 0.368 & 0.294 \\
    VS (DP)     & 0.932 & 0.804 & 0.672 & 1.000 & 0.161 & 0.606 & 0.643 \\
    \midrule
    TTDT (raw)    & 0.904 & 0.894 & 0.859 & 0.733 & 0.079 & 0.585 & 0.728 \\
    TTDT (DP)     & 0.922 & 0.860 & 0.766 & 0.933 & 0.142 & 0.658 & 0.728 \\
    \bottomrule
  \end{tabular}
\end{table}

\section{Discussion}
\label{sec:discussion}
Training-time domain transfer achieved diagnostic agreement comparable or superior to inference-time virtual staining without real unstained training data. Unlike domain adaptation~\cite{guan2022domain}, which typically requires target domain data to train the task model, our diagnostic model receives none at any stage, making the transfer zero-shot. The generative model needs paired data, but once trained it enables the diagnostic model training at scale from vast source domain archives. The viability of this paradigm rests on the hypothesis that fewer paired samples are needed to train a sufficient generative model than real target samples to train a diagnostic model from scratch. As in sim-to-real learning, synthetic training data need not perfectly reproduce the target domain~\cite{tobin2017domain}, only capture sufficient structure for the downstream task. 

Pixel-wise metrics such as SSIM or PSNR are sensitive to misalignment and may not reflect the preservation of diagnostically relevant features. Thus, rather than evaluating visual similarity alone, we focus on diagnostic utility of synthetic images as training data. Across both tissue processing stages, the generator-free approach showed higher agreement with the H\&E benchmark than virtual staining did, suggesting that avoiding inference-time generation may improve diagnostic consistency. 
Limitations of the study include data from a single laboratory and reference grading from the original collection; we therefore focused on concordance with the H\&E benchmark AI. Closing the remaining sim-to-real gap through larger datasets or improved generators remains future work.
In conclusion, training-time domain transfer removes generative AI from the diagnostic inference pipeline and enables zero-shot deployment on emerging modalities, where large training datasets are infeasible to collect. 

\section{Compliance with Ethical Standards}
\label{sec:ethics}
The study was approved by the Swedish Ethical Review Authority (2019-05220, 2025-07557-02).

\section{Acknowledgments}
Funding: SciLifeLab \& Wallenberg DDLS (KAW 2024.0159), Cancer Research KI. Computing: NAISS (funded by the Swedish Research Council), Berzelius (funded by KAW). 
Disclosures: A.C. received honoraria from Astellas Pharma;
A.C., N.M., K.K., and M.E. hold shares in Clinsight AB; M.E. holds shares and patents in A3P Biomedical AB.
 
\bibliographystyle{IEEEbib}
\bibliography{strings,refs}

\end{document}